\documentclass{article}

\usepackage[font=footnotesize,labelformat=simple]{subcaption}
\usepackage{tikz}
\usepackage{listofitems} 
\usetikzlibrary{arrows.meta} 
\usepackage[outline]{contour} 
\contourlength{1.4pt}

\tikzset{>=latex} 
\usepackage{xcolor}
\colorlet{myred}{red!80!black}
\colorlet{myblue}{blue!80!black}
\colorlet{mygreen}{green!60!black}
\colorlet{myorange}{orange!70!red!60!black}
\colorlet{mydarkred}{red!30!black}
\colorlet{mydarkblue}{blue!40!black}
\colorlet{mydarkgreen}{green!30!black}
\tikzstyle{node}=[thick,circle,draw=myblue,minimum size=22,inner sep=0.5,outer sep=0.6]
\tikzstyle{node in}=[node,green!20!black,draw=mygreen!30!black,fill=mygreen!25]
\tikzstyle{node hidden}=[node,blue!20!black,draw=myblue!30!black,fill=myblue!20]
\tikzstyle{node convol}=[node,orange!20!black,draw=myorange!30!black,fill=myorange!20]
\tikzstyle{node out}=[node,red!20!black,draw=myred!30!black,fill=myred!20]
\tikzstyle{connect}=[thick,mydarkblue] 
\tikzstyle{connect arrow}=[-{Latex[length=4,width=3.5]},thick,mydarkblue,shorten <=0.5,shorten >=1]
\tikzset{ 
  node 1/.style={node in},
  node 2/.style={node hidden},
  node 3/.style={node out},
}
\usepackage{microtype}
\usepackage{graphicx}
\usepackage{booktabs} 

\usepackage{hyperref}

\usepackage[accepted]{icml2023}

\usepackage{amsmath}
\usepackage{amssymb}
\usepackage{mathtools}
\usepackage{amsthm}

\usepackage[capitalize,noabbrev]{cleveref}

\theoremstyle{plain}

\theoremstyle{definition}

\theoremstyle{remark}

\usepackage[textsize=tiny]{todonotes}

\newcommand*{\selfattentiongraph}{
    \begin{tikzpicture}
      \def\yshift{0.4} 
      \def\xshift{0.4} 
    
      \def\NI{7} 
    
        \foreach \j [evaluate={\indexj=(\j<\NI?int(\j):"n");}] in {1,...,\NI}{
                \node[node in] (NI-\j) at (360/\NI * \j:2.5cm) {$x_{\indexj}$};
            }
        \foreach \j [evaluate={\indexj=(\j<\NI?int(\j):"n");}] in {1,...,\NI}{
    
            \foreach \i [evaluate={\indexi=(\i<\NI?int(\i):"n");}] in {1,...,\NI}{     
                \ifnum\i=1 
                    \node[node hidden] (NO-\i) at (360/\NI * \i:2.5cm) {$x_{\i}$};
                    \draw[connect] (NI-\j) -- (NI-\i);
                \else
                    \draw[connect, myblue!20] (NI-\j) -- (NI-\i);
                \fi
            }
        }
    \end{tikzpicture}
}

\newcommand*{\crossattentiongraph}{

    \begin{tikzpicture}[x=3.7cm,y=1.6cm]
      \def\NI{5} 
      \def\NO{4} 
      \def\yshift{0.4} 
      
      \foreach \i [evaluate={\c=int(\i==\NI); \y=\NI/2-\i-\c*\yshift; \index=(\i<\NI?int(\i):"n");}]
                  in {1,...,\NI}{ 
        \node[node in,outer sep=0.6] (NI-\i) at (0,\y) {$x_{\index}$};
      }
      \foreach \i [evaluate={\c=int(\i==\NO); \y=\NO/2-\i-\c*\yshift; \index=(\i<\NO?int(\i):"m");}]
        in {\NO,...,1}{ 
        \ifnum\i=1 
          \node[node hidden]
            (NO-\i) at (1,\y) {$x_{\index}'$};
          \foreach \j [evaluate={\index=(\j<\NI?int(\j):"n");}] in {1,...,\NI}{ 
            \draw[connect,white,line width=1.2] (NI-\j) -- (NO-\i);
            \draw[connect] (NI-\j) -- (NO-\i)
              node[pos=0.50] {\contour{white}{}};
          }
        \else 
          \node[node,blue!20!black!80,draw=myblue!20,fill=myblue!5]
            (NO-\i) at (1,\y) {$x_{\index}'$};
          \foreach \j in {1,...,\NI}{ 
            \draw[connect,myblue!20] (NI-\j) -- (NO-\i);
          }
        \fi
      }
      
      \path (NI-\NI) --++ (0,1+\yshift) node[midway,scale=1.2] {$\vdots$};
      \path (NO-\NO) --++ (0,1+\yshift) node[midway,scale=1.2] {$\vdots$};
    \end{tikzpicture}
}

\newcommand*{\mchngraph}{

    \begin{tikzpicture}[x=3.7cm,y=1.6cm]
      \def\NI{5} 
      \def\NO{4} 
      \def\yshift{0.4} 
      
      \foreach \i [evaluate={\c=int(\i==\NI); \y=\NI/2-\i-\c*\yshift; \index=(\i<\NI?int(\i):"n");}]
                  in {1,...,\NI}{ 
        \node[node in,outer sep=0.6] (NI-\i) at (0,\y) {$x_{\index}$};
      }
      \foreach \i [evaluate={\c=int(\i==\NO); \y=\NO/2-\i-\c*\yshift; \index=(\i<\NO?int(\i):"m");}]
        in {\NO,...,1}{ 
        \ifnum\i=1 
          \node[node,blue!20!black!80,draw=black!60,fill=black!20]
            (NO-\i) at (1,\y) {$z_{\index}$};
          \foreach \j [evaluate={\index=(\j<\NI?int(\j):"n");}] in {1,...,\NI}{ 
            \draw[connect,white,line width=1.2] (NI-\j) -- (NO-\i);
            \draw[connect] (NI-\j) -- (NO-\i)
              node[pos=0.50] {\contour{white}{}};
          }
        \else 
          \node[node,blue!20!black!80,draw=black!20,fill=black!5]
            (NO-\i) at (1,\y) {$z_{\index}$};
          \foreach \j in {1,...,\NI}{ 
            \draw[connect,myblue!20] (NI-\j) -- (NO-\i);
          }
        \fi
      }
      
      \path (NI-\NI) --++ (0,1+\yshift) node[midway,scale=1.2] {$\vdots$};
      \path (NO-\NO) --++ (0,1+\yshift) node[midway,scale=1.2] {$\vdots$};
    \end{tikzpicture}
}

\newcommand*{\slotattentiongraph}{

\begin{tikzpicture}[x=3.7cm,y=1.6cm]
      \def\NI{5} 
      \def\NO{4} 
      \def\yshift{0.4} 
      
      \foreach \i [evaluate={\c=int(\i==\NI); \y=\NI/2-\i-\c*\yshift; \index=(\i<\NI?int(\i):"n");}]
                  in {1,...,\NI}{ 
        \ifnum\i=1 
            \node[node in,outer sep=0.6] (NI-\i) at (0,\y) {$x_{\index}$};
        \else
            \node[node,blue!20!black!80,draw=mygreen!20,fill=mygreen!5] (NI-\i) at (0,\y) {$x_{\index}$};
        \fi
      }
      \foreach \i [evaluate={\c=int(\i==\NO); \y=\NO/2-\i-\c*\yshift; \indexi=(\i<\NO?int(\i):"m");}] in {\NO,...,1}{ 
              \foreach \j [evaluate={\indexK=(\j<\NI?int(\j):"n");}] in {1,...,\NI}{ 
                    \node[node,blue!20!black!80,draw=black!60,fill=black!20]
                (NO-\i) at (1,\y) {$z_{\indexi}$};
            \ifnum\j=1 
                \draw[connect,white,line width=1.2] (NI-\j) -- (NO-\i);
                \draw[connect] (NI-\j) -- (NO-\i);
                  node[pos=0.50] {\contour{white}{}};
            \else 
                \draw[connect,myblue!20] (NI-\j) -- (NO-\i);
        \fi
        }
      }
      
      \path (NI-\NI) --++ (0,1+\yshift) node[midway,scale=1.2] {$\vdots$};
      \path (NO-\NO) --++ (0,1+\yshift) node[midway,scale=1.2] {$\vdots$};
    \end{tikzpicture}
}

\newcommand*{\blockslotattentiongraph}{

\begin{tikzpicture}
[x=3.7cm,y=1.6cm]
      \def\NI{5} 
      \def\NO{4} 
      \def\NM{5} 

      \def\yshift{0.4} 

      \foreach \i [evaluate={\c=int(\i==\NI); \y=\NI/2-\i-\c*\yshift; \index=(\i<\NI?int(\i):"n");}]
                  in {1,...,\NI}{ 
        \ifnum\i=1 
            \node[node in,outer sep=0.6] (NI-\i) at (0,\y) {$x_{\index}$};
        \else
            \node[node,blue!20!black!80,draw=mygreen!20,fill=mygreen!5] (NI-\i) at (0,\y) {$x_{\index}$};
        \fi
      }
      \foreach \i [evaluate={\c=int(\i==\NO); \y=\NO/2-\i-\c*\yshift; \indexi=(\i<\NO?int(\i):"m");}] in {\NO,...,1}{ 
              \foreach \j [evaluate={\indexK=(\j<\NI?int(\j):"n");}] in {1,...,\NI}{ 
                    \node[node,blue!20!black!80,draw=black!60,fill=black!20]
                (NO-\i) at (1,\y) {$z_{\indexi}$};
            \ifnum\j=1 
                \draw[connect,white,line width=1.2] (NI-\j) -- (NO-\i);
                \draw[connect] (NI-\j) -- (NO-\i);
                  node[pos=0.50] {\contour{white}{}};
            \else 
                \draw[connect,myblue!20] (NI-\j) -- (NO-\i);
        \fi
        }
      }
      \foreach \i [evaluate={\c=int(\i==\NI); \y=\NM/2-\i-\c*\yshift; \index=(\i<\NM?int(\i):"l");}]
                  in {1,...,\NM}{ 
          \node[node hidden](NM-\i) at (2,\y) {$m_{\index}$};
        }
      \foreach \i [evaluate={\c=int(\i==\NM); \y=\NM/2-\i-\c*\yshift; \indexi=(\i<\NM?int(\i):"l");}] in {\NM,...,1}{ 
              \foreach \j [evaluate={\indexK=(\j<\NO?int(\j):"n");}] in {1,...,\NO}{ 
            \ifnum\j=1 
                \draw[connect,white,line width=1.2] (NO-\j) -- (NM-\i);
                \draw[connect] (NO-\j) -- (NM-\i);
                  node[pos=0.50] {\contour{white}{}};
            \else 
                \draw[connect,myblue!20] (NO-\j) -- (NM-\i);
        \fi
        }
      }
      
      \path (NI-\NI) --++ (0,1+\yshift) node[midway,scale=1.2] {$\vdots$};
      \path (NO-\NO) --++ (0,1+\yshift) node[midway,scale=1.2] {$\vdots$};
      \path (NM-\NM) --++ (0,1+\yshift) node[midway,scale=1.2] {$\vdots$};

    \end{tikzpicture}
}

\icmltitlerunning{Attention: Marginal Probabiliy is All You Need?}

\begin{document}

\twocolumn[
\icmltitle{Attention: Marginal Probability is All You Need?}



\icmlsetsymbol{equal}{*}

\begin{icmlauthorlist}
\icmlauthor{Ryan Singh}{sussex}
\icmlauthor{Christopher L. Buckley}{sussex}
\end{icmlauthorlist}

\icmlaffiliation{sussex}{School of Engineering and Informatics, University of Sussex}

\icmlcorrespondingauthor{Ryan Singh}{rs773@sussex.ac.uk}

\icmlkeywords{Machine Learning, ICML}

\vskip 0.3in
]



\printAffiliationsAndNotice{}  

\begin{abstract}
Attention mechanisms are a central property of cognitive systems allowing them to selectively deploy cognitive resources in a flexible manner.   Attention has been long studied in the neurosciences and there are numerous phenomenological models that try to capture its core properties.  Recently attentional mechanisms have become a dominating architectural choice of machine learning and are the central innovation of Transformers.  The dominant intuition and formalism underlying their development has drawn on ideas of keys and queries in database management systems.  In this work, we propose an alternative Bayesian foundation for attentional mechanisms and show how this unifies different attentional architectures in machine learning. This formulation allows to to identify commonality across different attention ML architectures as well as suggest a bridge to those developed in neuroscience. 
We hope this work will guide more sophisticated intuitions into the key properties of attention architectures as well suggest new ones. 
\end{abstract}

\section{Introduction}

Designing neural network architectures with favourable inductive biases lies behind many recent successes in Deep Learning \cite{baxter_model_2000}. In particular, the attention mechanism has  allowed language models to achieve human like generation abilities previously thought impossible \cite{vaswani_attention_2017}. The success of the attention mechanism as a domain agnostic architecture has prompted it to be adopted across a huge range of tasks and domains notably reaching state-of-the-art performance in visual reasoning and segmentation tasks \cite{dosovitskiy_image_2021, wang_image_2022}. 

Despite it's success, the role of the attention mechanism remains poorly understood. Indeed, it is unclear to what extent it relates to theories of cognitive attention which inspired it \cite{lindsay_attention_2020}. Here, we aim to provide a parsimonious description grounded in principles of probabilistic inference. This Bayesian perspective provides both a principled method for specifying prior beliefs and reasoning explicitly about the role of the attention variables. Further, understanding the fundamental computation permits us a unified description of different attention mechanisms in the literature. This proceeds in two parts.




First, we show that `soft' attention mechanisms (e.g. self-attention, cross-attention, graph attention, which we call \textit{transformer attention} herafter) can be understood probabilistically as taking an expectation over possible connectivity structures, providing an interesting link between softmax-based attention and marginal likelihood.

Second, we extend the uncertainty over connectivity to a bayesian setting which, in turn, provides a theoretical grounding for iterative attention mechanisms (slot-attention, perciever and block-slot attention) \cite{locatello_object-centric_2020, singh_neural_2022, jaegle_perceiver_2021} and Modern Continuous Hopfield Networks \cite{ramsauer_hopfield_2021}. 

Additionally, we apply iterative attention to Predictive Coding Networks, 
 an influential theory in computational neuroscience, creating a new theoretical bridge between machine learning and cognitive science.

\begin{equation*}
    \begin{split}
        Attention(Q, K, V) &= \overbrace{softmax(\frac{QW_{Q}W_K^TK^T}{\sqrt{d_k}})}^\text{$p(E \mid Q, K)$}V \\
        &= \mathbb{E}_{p(E\mid Q, K)}[V]
    \end{split}
\end{equation*}

A key observation is that the attention matrix can be seen as the posterior distribution over an adjacency structure, $E$, and the full mechanism as computing an expectation of the value function $V(X)$ over the posterior beliefs about the possible relationships that exist between key and query.

This formalism provides an alternate Bayesian theoretical framing within which to understand attention models, which contrasts with the original framing in terms of  database management systems and data retrieval, providing a unifying framework to describe different attention architectures. Describing their difference only in terms of their edge relationships supporting more effective analysis and development of new architectures. Additionally providing  a principled understanding of the difference between hard and soft attention models.

\textbf{Contributions}
\begin{itemize}
    \item A unifying probabilistic framework for understanding attention mechanisms.
    \item We show self-attention and cross-attention can be seen as computing a marginal likelihood over possible network structures.  
    \item We show that slot-attention, block-slot-attention and modern continuous hopfield networks can all be seen as collapsed variational inference, where the possible network structures form the collapsed variables.
    \item Provide a bridge to Bayesian conceptions of attention from computational neuroscience, through the lens of Predictive Coding Networks.
    \item Provide a framework for reasoning about hard attention, and efficient approximations to the attention mechanism.

\end{itemize}

\section{Related Work}
\textbf{Attention as bi-level optimisation}
Mapping feed-forward architecture to a minimisation step on a related energy function has been called unfolded optimisation \cite{frecon_bregman_2022}. Taking this perspective can lead to insights about the inductive biases involved for each architecture. It has been shown that the cross-attention mechanism can be viewed as an optimisation step on the energy function of a form of Hopfield Network \cite{ramsauer_hopfield_2021}, providing a link between attention and associative memory. Whilst \cite{yang_transformers_2022} extend this view to account for self-attention. Our framework distinguishes hopfield attention, which does not allow an arbritary value matrix, from the standard attention mechanisms. Whilst there remains a strong theoretical connection, it  places the Hopfield Energy as an instance of variational free energy, aligning more closely with iterative attention mechanisms such as slot-attention. 

\textbf{Relationship to gaussian mixture model}
Previous works that have taken a probabilistic perspective on the attention mechanism note the connection to inference in a gaussian mixture model \cite{gabbur_probabilistic_2021, nguyen_improving_2022, ding_attention_2020}. Indeed \cite{annabi_relationship_2022} directly show the connection between the Hopfield energy and the variational free energy of a gaussian mixture model. Although gaussian mixture models, a special case of the framework we present here, are enough to explain cross attention they do not capture slot or self-attention. Further our framework allows us to extend the structural inductive biases beyond what can be expressed in a gaussian mixture model and capture the relationship to hard attention.

\textbf{Latent alignment and hard attention}
Several attempts have been made to combine the benefits of soft (differentiability) and hard attention. Most approaches proceed by sampling, e.g., using the REINFORCE estimator \cite{deng_latent_2018} or a $topK$ approximation \cite{shankar_surprisingly_2018}. The one most similar to ours embeds the full forward-backward algorithm within a forward pass \cite{kim_structured_2017}, our approach differs by offering a parsimonious description in terms of marginalisation over an implicit graphical model.

\textbf{Collapsed Inference}
Collapsed variational inference has most notably been employed in topic modelling \cite{teh_collapsed_2006}. To our knowledge, linking collapsed inference to attention in deep learning is completely novel.

\section{Transformer Attention} \label{neural-attention}
\subsubsection{Attention as Expectation}
We begin by demonstrating transformer attention is best seen as an expectation over latent variables. In the case of self and cross-attention, the expectation of a neural network with respect to possible adjacency structures.  

Let $x=(x_1,..,x_n)$ be observed variables, $\phi$ be some set of latent variables, and $y$ a variable we need to predict. Given a latent variable model $p(y, x , \phi) = p(y \mid x, \phi)p(x, \phi)$,  where $p(y\mid x, \phi)$ is parameterised by some function $v(y, x, \phi)$ e.g. a neural network. 

Our goal is to find $p(y\mid x)$, however $\phi$ are unobserved so we calculate the marginal likelihood.

$$p(y \mid x) = \sum_\phi p(\phi \mid x)v(y, x, \phi)$$
Importantly, the softmax function is a natural representation for the posterior 
\begin{equation*}
    p(\phi \mid x) = \frac{p(x, \phi)}{\sum_{\phi} p(x,\phi)} \label{marginalised}
\end{equation*}
\begin{equation*} 
    p(\phi \mid x) = softmax(\ln p(x, \phi))
\end{equation*}
Hence, transformer attention can be seen as weighting $v(x, \phi)$ by the posterior distribution $p(\phi \mid x)$. 
\begin{equation}
\begin{split}
    p(y \mid x) &= \sum_{\phi}softmax(\ln p(x, \phi))v(y, x, \phi) \\
    &=\mathbb{E}_{p(\phi \mid x)}[v(y, x, \phi)]\label{attention-is-expectation}
\end{split}
\end{equation}

We claim \eqref{attention-is-expectation} is exactly the equation underlying self and cross-attention. To make a more direct connection, we present the specific generative models corresponding to them. The latent variables $\phi$ are identified as possible \textit{relationships}, or edges, between each of the observed variables $x$ (keys and queries). 

A natural formalism for modelling these graphical relationships is Markov Random Fields.

\subsubsection{Pairwise Markov Random Fields}
Given a set of random variables $X = (X_v)_{v\in V}$ with probability distribution $[p]$ and a graph $G = (V, E)$. The variables form a pairwise Markov random field (MRF) with respect to $G$ if the joint density function $P(X=x)=p(x)$ factorises as follows
$$p(x) = \frac{1}{Z}\exp \left( \sum_{v \in V} \psi_v +  \sum_{e \in E} \psi_e \right)$$
where $Z$ is the partition function $\psi_{v}(x_v)$ and $\psi_e=\psi_{u,v}(x_u, x_v)$ are known as the node and edge potentials respectively\footnote{See \cite{shah_learning_2021} for a precise definition.}.

Beyond the typical set-up, we add a structural prior $p(E)$ over the adjacency structure of the underlying graph.
\begin{equation*}
\begin{split}
   p(x, E) &= P(x\mid E)P(E) \\ 
    &= \frac{1}{Z}{p(E)\exp \left(\sum_{v \in V} \psi_v +  \sum_{e \in E} \psi_e\right)}
\end{split}
\end{equation*}

We briefly remark that \eqref{attention-is-expectation}  respects factorisation of $[p]$ in the following sense; if the distribution admits a factorisation with respect to the latent variables $p(x, \phi)=\prod_i f_i(x, \phi_i)$ and $v(x, \phi) =  \sum_{i} v_i(x, \phi_i)$ then (applying the linearity of expectation) we may write 
\begin{equation}
    \mathbb{E}_{p(\phi \mid x)}[v(x, \phi)] = \sum_i\mathbb{E}_{p(\phi_i \mid x)}[v_i] \label{attention-expectation-independence}
\end{equation}
Permitting each factor to be marginalised independently.

In the case of an MRF, such a factorisation is natural. If the distibution over edges factorises into local distributions $p(E)= \prod_i p(E_i)$ (using independence properties of the MRF) we can write $p(x, E) = \frac{1}{Z}\prod_i f_i(x, E_i)$ where each $f_i =  P(E_i) \exp\sum_{v\in V}\psi_v{\sum_{e \in E_i}\psi_e}$ is itself an unnormalised MRF.

To recover cross-attention and self-attention are such models with we need only specify a structural prior and potential functions.
\begin{figure}
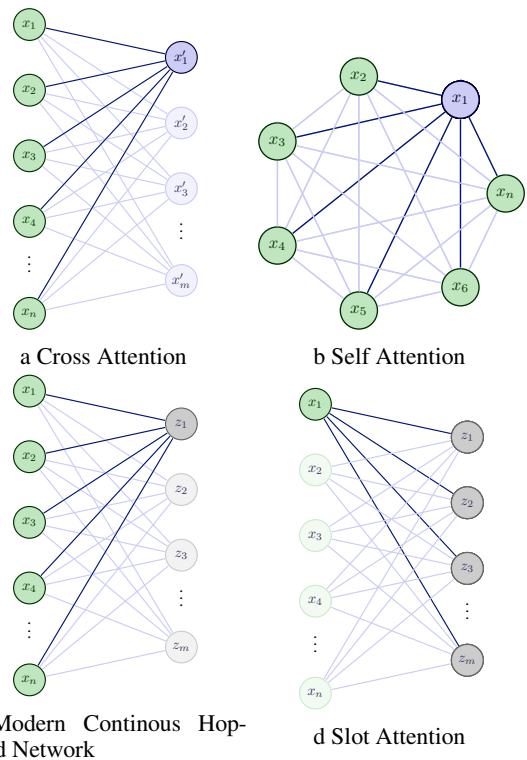

    \begin{subfigure}[t]{0.45\linewidth}
        \centering
        \resizebox{0.7\columnwidth}{!}{
            \crossattentiongraph
        }
        \caption{Cross Attention}
        \label{fig:cross-att}
    \end{subfigure}
    \begin{subfigure}[t]{0.45\linewidth}
    \centering
        \resizebox{\columnwidth}{!}{
            \selfattentiongraph
        }
        \caption{Self Attention}   
        \label{fig:self-att}
    \end{subfigure}

    \begin{subfigure}[c]{0.45\linewidth}
    \centering
        \resizebox{0.7\columnwidth}{!}{
            \mchngraph
        }
        \caption{Modern Continous Hopfield Network}   
        \label{fig:mchn}
    \end{subfigure}
    \begin{subfigure}[d]{0.45\linewidth}
    \centering
        \resizebox{0.7\columnwidth}{!}{
            \slotattentiongraph
        }
        \caption{Slot Attention}   
        \label{fig:slot-att}
    \end{subfigure}
\label{fig:subfig1.a.4}

\caption{
    Comparison of different attention modules in the literature, the highlighted edges is representative of the marginalisation being performed for the random variable $E_1$, in \ref{fig:cross-att} and \ref{fig:self-att} all nodes are observed, as opposed to \ref{fig:mchn} and \ref{fig:slot-att}, where there are latent nodes (indicated in grey).
}
\label{fig:my_label}
\end{figure}

\subsubsection{Cross Attention}
\begin{itemize}
    \item Key nodes  $K =(x_1,..,x_n)$
    \item Query nodes $Q = (x_1',...,x_m')$
    \item Structural prior $p(E) = \prod_{i=1}^m p(E_i)$, where $E_i \sim Uniform\{(x_1, x_i'),..,(x_n, x_i')\}$, such that each query node is uniformly likely to connect to each key node.
    \item Edge potentials $\psi(x_j, x_i')=x_i'^TW_{Q}^TW_{K}x_j$, in effect measuring the similarity of $x_j$ and $x_i'$ under a certain transformation.
    \item Value function $V_i(K, Q, E_i) = W_V x_{s(E_i)}$, a linear transformation applied to the node, $x_{s(E_i)}$, the start of the edge $E_i$.
\end{itemize}
Taking the posterior expectation in each of the factors defined in two \eqref{attention-expectation-independence} gives the standard cross- attention mechanism
$$\mathbb{E}_{p(E_i\mid Q, K)}[V_i] =  \sum_{j} softmax_j(x_i'^TW_{Q}^TW_{K}x_j)W_{V} x_j$$
$$\mathbb{E}_{p(E\mid Q, K)}[V] =  softmax(Q^TW_Q^TQ_KK)W_VK$$

\subsubsection{Self Attention}
\begin{itemize}
    \item Nodes  $K = Q = (x_1,..,x_n)$
    \item Structural prior $p(E) = \prod_{i=1}^n p(E_i^{\rightarrow})$, where $E_i^{\rightarrow} \sim Uniform\{(x_1, x_i),..,(x_n, x_i)\}$, such that each node is uniformly likely to connect to every other node.
    \item Edge potentials $\psi(k_j, k_i)=x_i^TW_{Q}^TW_{K}x_j$, in effect measuring the similarity of $x_j$ and $x_i'$ under a certain transformation.
    \item Value function $V_i(K, Q, E_i) = W_V x_{s(E_i)}$, a linear transformation applied to the node, $x_{s(E_i)}$, the start of the edge $E_i$.
\end{itemize}
Again, taking the posterior expectation in each of the factors defined in two \eqref{attention-expectation-independence} gives the standard self- attention mechanism
$$\mathbb{E}_{p(E_i\mid Q, K)}[V_i] =  \sum_{j} softmax_j(x_i^TW_{Q}^TW_{K}x_j)W_{V} x_j$$
$$\mathbb{E}_{p(E\mid Q, K)}[V] =  softmax(K^TW_Q^TW_KK)W_VK$$

\section{Iterative Attention}
We continue by extending attention to full Bayesian inference. In essence applying the attention trick, marginalisation of attention variables, to the variational free energy (a.k.a the ELBO).

Modern Continuous Hopfield Networks can be seen as a particular instance of this class of system, allowing us to reproduce the `hopfield attention' updates of \cite{ramsauer_hopfield_2021} within a probabilistic context. Under different structural priors we  recover other iterative attention models; slot-attention \cite{locatello_object-centric_2020}, block-slot attention \cite{singh_neural_2022} and Perciever \cite{jaegle_perceiver_2021}. Further, we showcase a specific advantage of bayesian attention, hard attention.

\subsubsection{Collapsed Inference}
We present a version of collapsed variational inference \cite{teh_collapsed_2006} showing how this results in a bayesian attention mechanism. The term attention mechanism is apt due to the surprising similarity in form between the variational updates \eqref{expected_attention} and neural attention mechanism \eqref{attention-is-expectation}. 

Our setting is the latent variable model $p(x,z, \phi)$, where $x$ are observed variables, and $z$, $\phi$, are latent variables. Typically  we wish to infer $z$ given $x$. 

Collapsed inference proceeds by marginalising out the extraneous latent variables $\phi$
\begin{equation}
    p(x, z) = \sum_{\phi} p(x,z, \phi) \label{marginalised}
\end{equation}

We define a recognition density $q(z) \sim N(z; \mu)$ and optimise the variational free energy with respect to the parameters, $\mu$, of this distribution.
$$\min_{\mu} F(x, \mu) = \mathbb{E}_q[\ln q_\mu(z) - \ln p(x,z)]$$
Under a typical  Laplace approximation, we can write the variational free energy as  $F \approx -\ln p(x, \mu)$ \footnote{See appendix for a more principled derivation taking account of higher order terms}.  Substituting in \eqref{marginalised} and taking the derivative with respect to the variational parameters yields,
$$F(x, \mu) = - \ln \sum_{\phi} p(x, \mu, \phi)$$
\begin{equation}
    \frac{\partial F}{\partial \mu} = - \frac{1}{\sum_{\phi} p(x, \mu, \phi)}\sum_{\phi} \frac{\partial}{\partial \mu}p(x, \mu, \phi) \label{general_attention}
\end{equation}

Which connects bayesian attention with the standard attention \eqref{attention-is-expectation}. To clarify this, we employ the log-derivative trick, substituting $p_\theta = e^{\ln p_\theta}$ and re-express \eqref{general_attention} in two ways:
\begin{equation}
    \frac{\partial F}{\partial \mu} = -\sum_{\phi} softmax_{\phi}(\ln p(x, \mu, \phi))\frac{\partial}{\partial \mu}\ln p(x, \mu, \phi) \label{softmax_derivative}
\end{equation}
\begin{equation}
    \frac{\partial F}{\partial \mu} = \mathbb{E}_{p(\phi \mid x, \mu)}[-\frac{\partial}{\partial \mu}\ln p(x, \mu, \phi)] \label{expected_attention}
\end{equation}

The first form reveals the softmax which is ubiquitous in all attention models. The second, suggests the variational update should be evaluated as the expectation of the  typical variational gradient (the term within the square brackets) with respect to the posterior over the parameters represented by the random variable $\phi$. 

In other words, bayesian attention is exactly the nueral attention mechanism applied iteratively, where the value function is the variational free energy gradient. We derive updates for a general MRF before again recovering (iterative) attention models in the literature  by specifying particular distributions.

\subsubsection{Free Energy of a marginalised MRF}

Recall the factorised MRF, $p(E)= \prod_i p(E_i)$.  $p(x, E) = \frac{1}{Z} \prod_i f_i(x, E_i)$ with each $f_i =  P(E_i)\exp\sum_{v\in V}\psi_v{\sum_{e \in E_i}\psi_e}$. Independence properties mean the  marginalisation necessary for collapsed inference can be simplified

\begin{equation*}
        \sum_E p(x, E) = \frac{1}{Z}\prod_i \sum_{E_i} f_i(x, E_i)
\end{equation*}

In an inference setting the nodes are partitioned into observed nodes, $x$, and latent nodes, $z$. The variational free energy \eqref{general_attention} and the associated forms of it's derivative can be expressed  
\begin{equation*}  
    F(x, \mu, \theta) = - \sum_i\ln \sum_{E_i} f_i(x, \mu, E_i) 
\end{equation*}
$$\frac{\partial F}{\partial \mu_j} =  -\sum_i \sum_{E_i} softmax(f_i(x, \mu, E_i))\frac{\partial f_i}{\partial \mu_j}$$

Similar to hard attention approaches, the random variable $E$  is an explicit alignment variable. However, unlike hard attention, we avoid inferring $E$ \textit{explicitly} using the collapsed inference approach outlined above.

\subsubsection{Quadratic Potentials and the convex concave procedure}
We follow \cite{ramsauer_hopfield_2021} in using the CCCP to derive a fixed point equation, which necessarily reduces the free energy. 

Assuming the node potentials are quadratic $\psi(x_i) = -\frac{1}{2}x_i^2$ and the edge potentials have the form $\psi(x_i, x_j) = x_iWx_j$.
\begin{equation}
    \mu_j^{*} =  \sum_i \sum_{E_i} softmax(g_i(x, \mu, E_i))\frac{\partial g_i}{\partial \mu_j} \label{fixed_point_attention}
\end{equation}
Where $g_i = \sum_{e \in E_i}\psi_e$.

By way of the CCCP \cite{yuille_concave-convex_2001}, this fixed point equation has the property $F(x, \mu_j^*, \theta) \leq F(x, \mu_j, \theta)$ with equality if and only if $\mu_j^*$ is a stationary point of $F$.

We follow the \ref{neural-attention} in specifying specific structural priors and potential functions to recover different iterative attention mechanisms.

\subsubsection{Hopfield-Style Cross Attention}
Let the observed $x =(x_1,..,x_n)$ and latent nodes $z=(z_1,..,z_m)$ have the following structural prior $p(E) = \prod_{i=1}^m p(E_i)$, where $E_i \sim Uniform\{(x_1, z_i),..,(x_n, z_i)\}$. And define edge potentials $\psi(x_j, z_i)=z_iQ^TKx_j$, Application of \eqref{fixed_point_attention}
$$\mu_i^{*} =  \sum_{j} softmax_j(\mu_iW_{Q}^TW_Kx_j)W_Q^T W_K x_j$$

When $\mu_i$ is initialised to some query $\xi$ the system
\cite{ramsauer_hopfield_2021} the fixed point update is given by $\mu_i^{*}(\xi) = \mathbb{E}_{p(E_i \mid x, \xi)}[W_Q^T W_Kx_{t(E_i)}]$. When the patterns $x$ are well separated, $\mu_i^*(\xi) \approx W_Q^T W_Kx_j$, where $W_Q^T W_Kx_j$ is the closest vector and hence can be used as an associative memory.

\subsubsection{Slot Attention}
Slot attention \cite{locatello_object-centric_2020} is an object centric learning module built on top of an iterative attention mechanism. Here we show this is a simple adjustment of the prior beliefs on our edge set.

With the same set of nodes and potentials, replace the prior over edges with 
$p(E) = \prod_{j=1}^n p(E_j)$, $E_j \sim Uniform\{(x_j, z_1),..,(x_j, z_m)\}$
\begin{equation*}
    \mu_i^{*} =  \sum_{j} softmax_i(\mu_iQ^TKx_j)Q^T K x_j
\end{equation*}

Whilst the original slot attention employed an RNN to aid the basic update shown here, the important feature is that the softmax is taken over the `slots', $\mu$. This forces competition between slots to account for the observed variables, forcing object centric representations. For example, if the observed variables $x$ are image patches, the slots are forced to cluster similar patches together in order increase the overall likelihood of said patches. The word cluster is accurate, in fact there is an exact equivalence between this mechanism and a step of EM on a gaussian mixture model.



\subsubsection{Block Slot Attention}
\cite{singh_neural_2022} suggest combining an associative memory ability with an object-centric slot-like ability and provide an iterative scheme for doing so, alternating between slot-attention and hopfield updates. 

Our framework permits us to flexibly combine different attention mechanisms through different latent graph structures, allowing us to derive a model informed version of block-slot attention. In this setting we have three sets of variables $X$, the observations, $Z$ the latent variables to be inferred and $M$ which are parameters.

Define the pairwise MRF $X = \{x_1,...,x_n\}$, $Z = \{z_1,...,z_m\}$ and $M = \{m_1,...,m_l\}$ with a prior over edges  $p(E) = \prod_{j=1}^m p(E_j)\prod_{k=1}^l p(\Tilde{E_k})$, $E_j \sim Uniform\{(x_j, z_1),..,(x_j, z_m)\}$, $\Tilde{E_k} \sim Uniform\{(z_1, m_k),..,(z_m, m_k)\}$, with edge potentials between $X$ and $Z$ given by $\psi(x_j, z_i)=z_iQ^TKx_j$ and between $Z$ and $M$, $\psi(z_i, m_k)=z_i \cdot m_k$

applying \eqref{fixed_point_attention} gives
\begin{equation*}
    \begin{split}
        \mu_i^{*} =  &\sum_{j} softmax_i(\mu_iQ^TKx_j)Q^T K x_j \\
                    & + \sum_{k} softmax_k(\mu_i \cdot m_k)m_k
    \end{split}
\end{equation*}

\begin{figure}
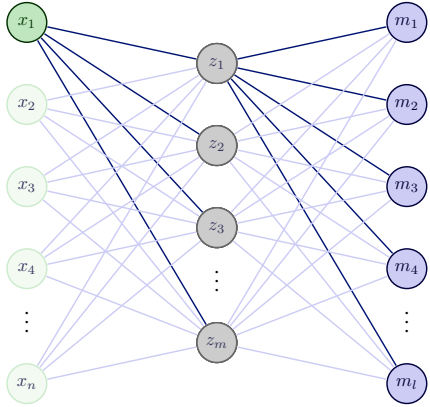

    \centering
        \resizebox{0.7\columnwidth}{!}{
        \blockslotattentiongraph
        }
    \caption{Block Slot Attention}
    \label{fig:my_label}
\end{figure}

In the original block-slot attention each slot $z_i$ is broken into blocks, where each block can access block-specific memories i.e. $z_i^{(b)}$ can has possible connections to memory nodes $\{m_k^{(b)}\}_{k\leq l}$. Allowing objects to be represented by slots which in turn disentangle features of each object in different blocks. We presented a single block version above, however it is easy to see that the update extends to the multiple block version 
applying \eqref{fixed_point_attention} gives
\begin{equation*}
    \begin{split}
        \mu_i^{*} =  &\sum_{j} softmax_i(\mu_iQ^TKx_j)Q^T K x_j \\
                    & + \sum_{k, b} softmax_k(\mu_i^{(b)} \cdot m_k^{(b)})m_k^{(b)}
    \end{split}
\end{equation*}

\section{Predictive Coding Networks}

Predictive Coding Networks (PCN) have emerged as an influential theory in computational neuroscience \cite{rao_predictive_1999, friston_predictive_2009, buckley_free_2017}. Building on theories of perception as inference and the Bayesian brain, PCNs perform approximate Bayesian inference by minimising the variational free energy which is manifested in the minimisation of local prediction errors. The continuous time dynamics at an individual neuron are given by
$$
    \frac{\partial \mathcal{F}}{\partial \mu_i} = -\sum_{\phi^{-}} k_{\phi}\epsilon_{\phi} + \sum_{\phi^{+}}k_{\phi}\epsilon_{\phi}w_{\phi}
$$
Where $\epsilon$ are prediction errors, $w$ represent synaptic strength and $k$ are node specific precisions representing uncertainty in the generative model \cite{millidge_theoretical_2022}.

A natural extension is to apply collapsed inference over the set of incoming and out going connection, i.e. a locally factorised prior over possible connectivity. In the notation of the previous section, we have an MRF with a hierarchical structure $Z = \{Z^{(0)}, ..., Z^{(l)}, ..., Z^{(N)} \}$ where the prior on edges factorises into layerwise $p(E^{(l)})=\{(z_i, z_j): (z_i, z_j) \in Z^{(l-1)}\times Z^{(l)}\}$ and potential functions $\phi(z_i, z_j)=\epsilon_{i,j}^2 = k_{j}(z_j - w_{i,j}z_i)^2$.
\begin{equation*}
    \begin{split}
          \frac{\partial \mathcal{F}}{\partial \mu_i} =  &-\sum_{\phi^{-}} softmax( {-\epsilon_\phi}^2)k_{\phi}\epsilon_{\phi} \\
                    & + \sum_{\phi^{+}} softmax( {-\epsilon_\phi}^2)k_{\phi}\epsilon_{\phi}w_{\phi}
    \end{split}
\end{equation*}

The resulting dynamics induce a ``normalisation" across prediction errors received by a neuron through the softmax function. This dovetails nicely with theories of attention as normalisation in psychology and neuroscience. In contrast previous predictive coding based theories of attention have focused on the precision terms, $k$, due to their ability to up and down regulate the impact of prediction errors \cite{feldman_attention_2010}. Here we see the softmax term can also perform this regulation, while also exhibiting the fast winner-takes-all dynamics that are associated with cognitive attention.

\subsection{Discussion}

In this section we will briefly discuss what can be gained from looking at the attention mechanism as a problem of inference.

\subsubsection{Hard Attention}


Recall \eqref{attention-is-expectation} neural attention may be viewed as calculating an expectation over latent variables $\mathbb{E}_{p(\phi \mid x)}[v(x, \phi)]$. Here the mechanism is `soft' because we weight multiple possibilities of attention variable $\phi$. Hard attention, on the other hand, proceeds with a single sample from $p(\phi \mid x)$. It has been argued this is more biological, more interpretable and has lower computational complexity. Previously the inferior performance of hard-attention has been attributed to it's hard to train, stochastic nature. However, our framing of soft attention as exact marginalisation offers an alternate explanation. Stochastic approximations (hard attention) will always suffer compared with exact marginalisation (soft attention). Further our framework provides a method for seamlessly interchanging hard and soft-attention. Since the distribution $p(\phi \mid x)$ a the categorical distribution, at any point (during training or inference) it is possible to implement hard attention by taking a single sample $\phi^*$ from $p(\phi \mid x)$ yielding $v(x, \phi^*)$. 

There are two issues with this approach to collapsing the attention distribution. First, the single sample will collapse any uncertainty, secondly calculation of $p(\phi \mid x)$, in order to sample, still incurs a quadratic penalty $O(n^2)$. However we can employ tools from probability theory to help us analyse the cost of sampling, and linear approximations to the attention distribution.

\subsubsection{Efficient Transformers}

Consider some distribution $q$ attempting to approximate $p(\phi \mid x)$ we can quantify the information loss with the relative entropy 
$$\mathcal{L}[p,q] \triangleq D_{KL}[q(\phi) \mid \mid p(\phi \mid x)] = H[q] + \mathbb{E}_q[p(\phi \mid x)]$$
In the hard attention approximation a single sample from $p$ is used as an approximation $\mathcal{L}[p,q] = -\ln p(\phi^* \mid x)$ and perhaps intuitively $\mathbb{E}[\mathcal{L}] = H[p]$ i.e. hard attention is a good approximation when the attention distribution is low-entropy which can be controlled by the temperature parameter (Appendix \ref{Temperature}).

Many of the efficient alternatives to attention, such as low-rank and linear approximations, can be cast as approximating $p(\phi \mid x)$ with $q(\phi \mid x)$ where calculating $q$ is less expensive than exact marginalisation. Estimating $\mathcal{L}$ could be used to quantify the relative information loss when using these alternatives. Another direction taken to reduce computational complexity of the attention mechanism is sparsification the attention matrix, which in our framework reduces to adjustments to the prior over edges (Appendix \ref{positional-encodings}).

\subsubsection{New Designs}

The main difference between the description presented and previous probabilistic descriptions is to view soft attention as a principled, exact, probabilistic calculation, with respect to an implicit probabilistic model, as opposed to an impoverished approximation. This leads to possibility of designing new attention mechanisms by altering the distribution that the mechanism marginalises over, either by adjusting the structural prior, or the potential functions. We hope this will enable new architectures to be designed in a principled manner.




\bibliography{main}
\bibliographystyle{icml2023}

\newpage
\appendix
\onecolumn

\end{document}